\documentclass[conference]{IEEEtran}
\IEEEoverridecommandlockouts

\usepackage{cite}
\usepackage{amsmath,amssymb,amsfonts}
\usepackage{algorithmic}
\usepackage{graphicx}
\usepackage{textcomp}
\usepackage{xcolor}
\usepackage{booktabs}
\usepackage{multirow}
\usepackage{url}
\def\BibTeX{{\rm B\kern-.05em{\sc i\kern-.025em b}\kern-.08em
    T\kern-.1667em\lower.7ex\hbox{E}\kern-.125emX}}
\begin{document}

\title{Agentic Knowledge Tracing: A Multi-Agent LLM Architecture for Stealth Assessment of Financial Literacy in Serious Games
}

\author{\IEEEauthorblockN{ Gabriel Machado Santos}
\IEEEauthorblockA{\textit{Federal University of Uberlândia} \\
\textit{Computer Science Faculty}\\
São Paulo, Brazil \\
gabrielmsantos@gmail.com}
\and
\IEEEauthorblockN{Rita Maria Silva Julia}
\IEEEauthorblockA{\textit{Federal University of Uberlândia} \\
\textit{Computer Science Faculty}\\
Uberlândia, Brazil \\
rita@ufu.br}
\and
\IEEEauthorblockN{Marcelo Zanchetta do Nascimento}
\IEEEauthorblockA{\textit{Federal University of Uberlândia} \\
\textit{Computer Science Faculty}\\
Uberlândia, Brazil \\
marcelo.nascimento@ufu.br}
}

\IEEEoverridecommandlockouts

\IEEEpubid{\makebox[\columnwidth]{979-8-3315-9476-3/26/\$31.00~\textcopyright~2026 IEEE \hfill}
\hspace{\columnsep}\makebox[\columnwidth]{}}

\maketitle

\IEEEpubidadjcol

\begingroup
\renewcommand\thefootnote{}
\footnotetext{%
Accepted for publication in the IEEE Conference on Games (CoG) 2026.

\copyright~2026 IEEE. Personal use of this material is permitted.
Permission from IEEE must be obtained for all other uses, in any current
or future media, including reprinting/republishing this material for
advertising or promotional purposes, creating new collective works, for
resale or redistribution to servers or lists, or reuse of any copyrighted
component of this work in other works.}
\addtocounter{footnote}{-1}
\endgroup

\begin{abstract}
Assessing financial literacy during gameplay without disrupting the learning experience remains a key challenge in serious games for education. We present the Agentic BKT pipeline, a multi-agent large language model architecture for stealth assessment of financial competencies from open-ended gameplay events. The pipeline processes events from a 2D platformer serious game aligned with the OECD/INFE financial literacy framework through four phases: (1)~the game captures every player decision as a structured event log; (2)~an LLM event classifier labels each action on a four-point rubric validated against three domain experts (Fleiss kappa = 0.624, substantial agreement); (3)~four domain-specific agents specializing in risk mitigation, investing, spending, and credit management perform session-level reasoning over behavioral trajectories, feeding per-competency Bayesian Knowledge Tracing that estimates mastery within each domain; and (4)~an expert judge agent synthesizes the domain-level estimates into an overall mastery score. Evaluated with 193 K-12 participants across 264 game sessions, the Agentic BKT pipeline yields mastery estimates significantly correlated with learning gain (r = 0.276, p = 0.0001) and post-test scores (r = 0.333, p \textless{} 0.0001) while showing no correlation with pre-test scores, providing both convergent and discriminant validity. The multi-agent approach approximately triples the predictive validity of a single-LLM baseline (r = 0.095, not significant) in this study, demonstrating that domain decomposition and session-level reasoning play a central role in capturing the multidimensional nature of financial literacy from gameplay.
\end{abstract}


\section{Introduction}

Financial literacy---the knowledge, skills, and attitudes required for effective financial decision-making---is increasingly recognized as essential for modern economic participation~\cite{oecd2020infe}. However, OECD assessments show that 18\% of 15-year-olds lack basic financial proficiency, with socio-economic background strongly influencing outcomes, and many adolescents already engage in financial activities such as online purchases and bank account usage~\cite{oecd2024pisa}. Therefore, scalable and engaging educational interventions are urgently needed.

Serious games have emerged as promising tools for financial literacy education by providing interactive and experiential learning environments~\cite{amagir2018review, aprea2018instructional}. Prior studies show that such games can improve financial knowledge and behaviors~\cite{pfandler2021happy, platz2025financial}. However, assessing learning during gameplay remains challenging. Most systems either rely on intrusive post-game quizzes or omit assessment entirely, limiting insight into player learning~\cite{shute2009melding}. Stealth assessment addresses this issue by embedding assessment directly into gameplay~\cite{shute2011stealth, shute2013measuring}, but its application to financial literacy games remains limited.

A second challenge concerns methodology. Traditional knowledge tracing models such as Bayesian Knowledge Tracing (BKT)~\cite{corbett1995knowledge} assume structured right/wrong interactions tied to discrete skills. In open-ended games, however, player behavior is contextual and temporal: repeated gambling after losses reflects a different competence profile than abandoning gambling after negative outcomes, despite similar event types. Recent work suggests that large language models (LLMs) can support knowledge tracing and educational assessment through natural language understanding~\cite{schmucker2024towards, scarlatos2025dialogue, llmagents2025education}, but no prior work has applied multi-agent LLM systems to stealth assessment of financial literacy in serious games.

To address these challenges, we present a 2D platformer serious game aligned with the OECD/INFE financial competency framework~\cite{oecd2020infe} together with a novel \textbf{Agentic BKT pipeline}. The game records structured gameplay events, including purchases, investments, gambling, credit usage, and debt-related decisions, along with full game-state context. The proposed pipeline consists of four stages: (1) event logging during gameplay; (2) LLM-based event classification using a four-level rubric (POOR, FAIR, GOOD, EXCELLENT), validated against expert annotations; (3) four domain-specific LLM agents specialized in risk mitigation, investing, spending, and credit/debt, each combined with Bayesian Knowledge Tracing to model temporal competency development; and (4) a judge agent that synthesizes domain-level estimates into an overall mastery score. We evaluate the approach with 193 K-12 participants across 264 gameplay sessions, comparing it against a single-LLM baseline and validating mastery estimates against pre- and post-test scores.

Our contributions are:
\begin{enumerate}
    \item A novel multi-agent LLM framework for \textbf{stealth assessment} in financial literacy games, combining event classification, domain-specific reasoning, and interpretable knowledge tracing.
    
    \item Empirical evidence that the proposed approach produces mastery estimates significantly correlated with learning gain ($r = 0.276$, $p = 0.0001$) and post-test scores ($r = 0.333$, $p < 0.0001$), while remaining uncorrelated with pre-test scores.
    
    \item Evidence that \textbf{competency decomposition matters}: the multi-agent architecture approximately triples predictive validity in this evaluation compared to a single-LLM baseline ($r = 0.095$, $p = 0.19$).
    
    \item A reusable framework for aligning open-ended gameplay events with established competency standards such as OECD/INFE.
\end{enumerate}

The remainder of the paper investigates three research questions: \textbf{RQ1}, how well does the LLM event classifier agree with human domain experts; \textbf{RQ2}, to what extent do mastery estimates produced by the Agentic BKT pipeline correlate with external learning outcomes; and \textbf{RQ3}, how much of this predictive validity is attributable to domain decomposition versus a monolithic single-LLM baseline.

\section{Related Work}

The OECD/INFE framework defines core competencies---including knowledge of financial products, risk awareness, and responsible credit use---that serve as a reference standard for financial literacy curricula worldwide~\cite{oecd2020infe}. Translating these competencies into engaging learning experiences has motivated a growing body of serious games for financial literacy education. The Happy Life Game~\cite{pfandler2021happy} immerses players in life-stage financial decisions, while NOVA Financial Lab~\cite{corbin2022nova} targets behavioral biases in spending and saving through behavioral-science-based mini-games. FinCraft~\cite{rasco2020fincraft} proposes a personalised persuasive gaming platform, and Moonshot~\cite{platz2025financial} integrates instructional design principles into secondary education. Despite their promise, these systems generally rely on external pre/post-tests or self-report instruments, offering limited visibility into how financial competence develops \textit{during} gameplay. Bridging this gap between gameplay and assessment is the primary motivation for our work.

The challenge of embedding assessment into interactive learning environments has been addressed most directly by the stealth assessment paradigm~\cite{shute2011stealth, shute2013measuring}, which operationalizes ECD by defining competency, evidence, and task models that map observable in-game behaviors to latent competencies. BKT~\cite{corbett1995knowledge} provides a principled probabilistic framework for modeling skill acquisition, but its assumption of discrete, binary-graded responses limits applicability to the sequential and unstructured nature of gameplay. Deep learning extensions, including Deep Knowledge Tracing (DKT)~\cite{piech2015deep} and attention-based models such as AKT~\cite{ghosh2020context}, offer greater flexibility but require large labeled datasets rarely available in serious game deployments. More recently, LLM-based KT approaches have shown strong performance: LLMKT~\cite{scarlatos2025dialogue} outperformed traditional KT models in tutoring dialogues, NTKT~\cite{norris2025ntkt} reframed KT as next-token prediction, and 2T-KT~\cite{li2025twotkt} used LLM ``teacher thinking'' to predict student performance on unseen concepts. However, none has explored stealth assessment for financial literacy games.

A parallel line of work has explored multi-agent LLM architectures for educational tasks. Recent surveys describe systems in which specialized agents---such as administrators, evaluators, and critics---collaborate to estimate student knowledge states~\cite{llmagents2025education}. Beyond education, the LLM-as-a-judge paradigm has achieved over 80\% agreement with human preferences in open-ended evaluation tasks~\cite{zheng2023judging}, supporting scalable AI-driven assessment. Our work combines these directions into an integrated pipeline for stealth assessment in serious games, unifying multi-agent LLM event classification, domain-specific knowledge tracing, and expert-judge synthesis. Structured around the OECD/INFE framework and validated in a real educational setting, our approach provides the first empirical evidence that such an architecture can produce mastery estimates with meaningful predictive validity for financial literacy learning outcomes.

\section{Game Design}

The game used in this study is a 2D side-scrolling platformer built in Unity and deployed via WebGL, targeting K-12 students. Players begin each session with a fixed balance and must maximize their total wealth within a configurable time limit (10, 15, or 20 minutes). The core gameplay loop combines exploration and coin collection with financial decision-making: players encounter interactive elements that allow them to spend, invest, save, gamble, and use credit. This design ensures that every session produces a rich sequence of financially relevant decisions without requiring explicit financial instruction.

\begin{figure}[t]
\centering
\includegraphics[width=\columnwidth]{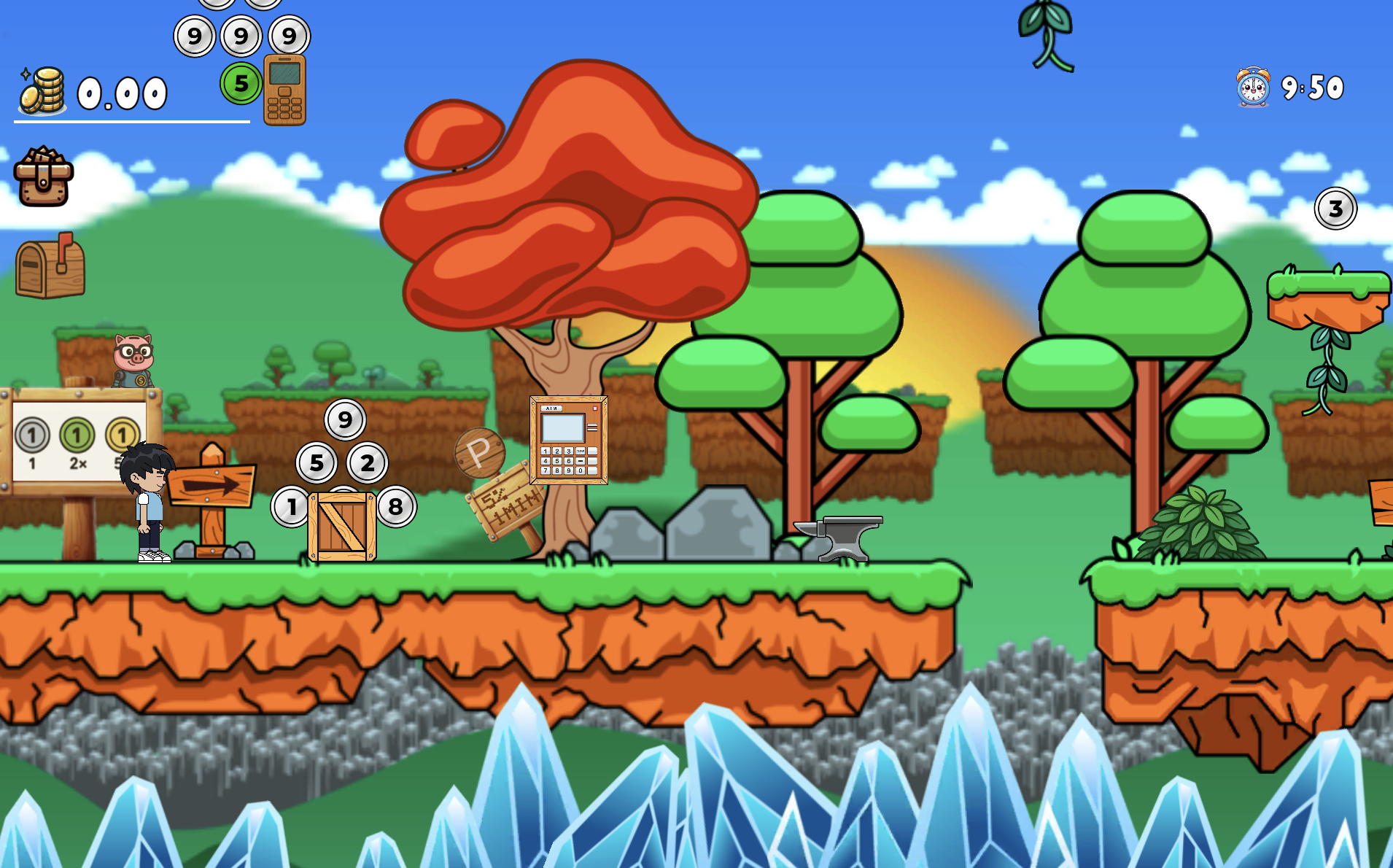}
\caption{Game overview showing the platformer environment. The HUD displays the player's current balance (top left), inventory and calculator access (top center), and remaining time (top right). Interactive objects include an investment sign (left), a deposit box, and an ATM.}
\label{fig:gameplay}
\end{figure}

\begin{figure}[t]
\centering
\includegraphics[width=\columnwidth]{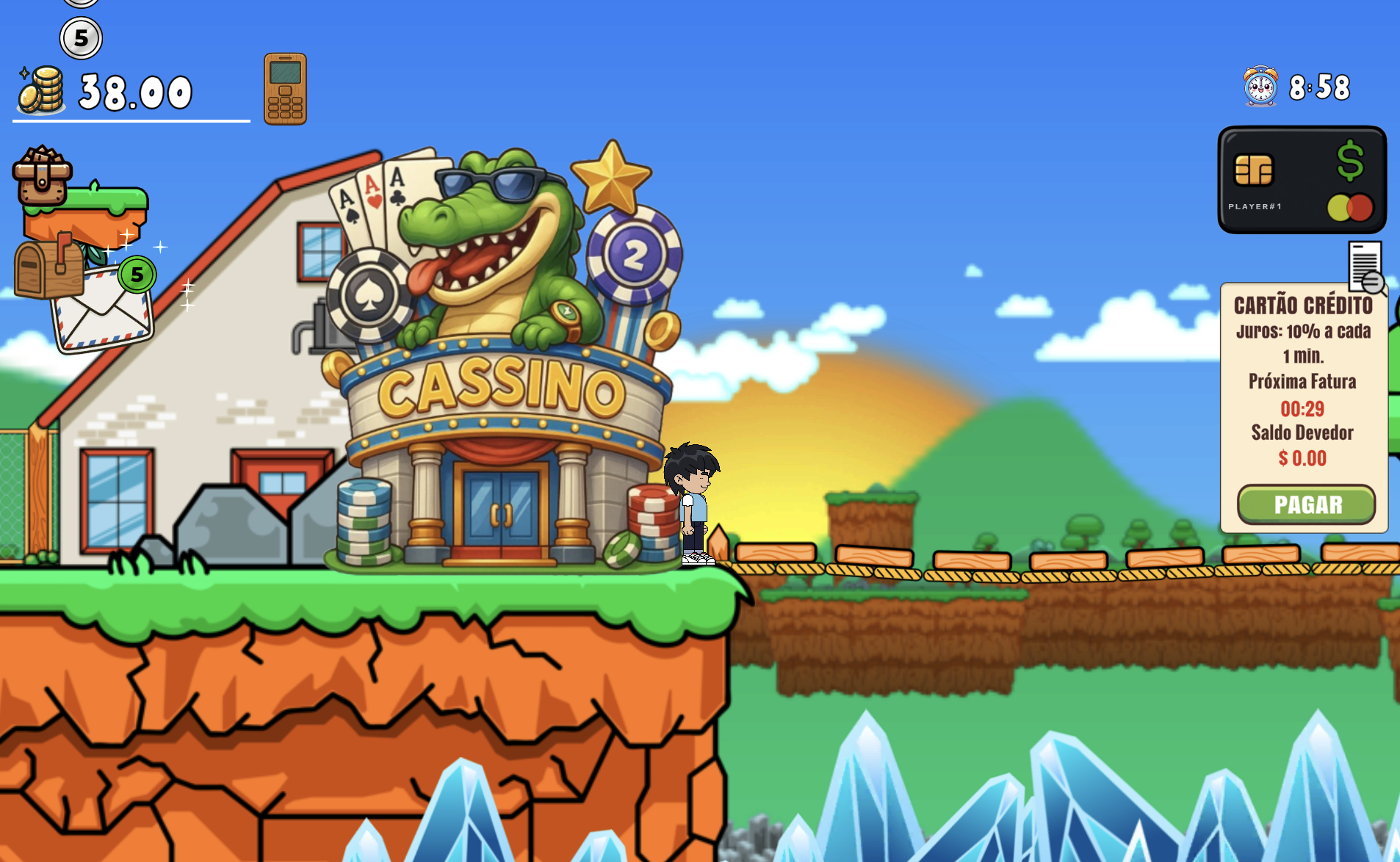}
\caption{Casino area and credit card interface. The player stands in front of the casino entrance, while the credit card panel (right) shows the card holder's name, interest rate (10\% per minute), next billing cycle, and outstanding debt with a payment button.}
\label{fig:casino}
\end{figure}

The game mechanics are explicitly aligned with the OECD/INFE financial literacy framework~\cite{oecd2020infe}, as shown in Table~\ref{tab:mechanics}. Purchasing items with limited funds exercises competencies in Money \& Transactions and Planning. Short- and long-term investments, where players choose among different interest rates and durations, engage Risk \& Reward alongside Planning. A casino with slot machines implements a configurable Return to Player (RTP) percentage, providing direct experience with negative expected value---a core concept in the Risk \& Reward domain. A credit card system enables spending beyond the player's balance, with interest accruing at a fixed rate per minute and periodic billing cycles, exercising Planning and Financial Landscape competencies. Finally, a bank account allows players to create savings accounts, touching Planning \& Managing. This alignment ensures that the game produces behaviorally diverse events that map onto recognized financial competency dimensions.

\begin{table}[t]
\centering
\caption{Mapping of game mechanics to OECD/INFE domains.}
\label{tab:mechanics}
\renewcommand{\arraystretch}{1.15}
\begin{tabular}{p{1.6cm}p{2.5cm}p{3.2cm}}
\hline
\textbf{Mechanic} & \textbf{OECD/INFE Domain} & \textbf{Description} \\
\hline
Purchasing items & Money \& Transactions; Planning & Spending decisions with limited balance \\
Investments & Risk \& Reward; Planning & Choose interest rates and durations \\
Casino / slots & Risk \& Reward & Gambling with negative expected value (RTP) \\
Credit card & Planning; Financial Landscape & Debt accumulation, interest accrual, payments \\
Bank account & Planning \& Managing & Saving, account creation \\
\hline
\end{tabular}
\end{table}

The game logs every player action as a structured text event containing the following fields: \texttt{event\_type}, \texttt{description}, \texttt{amount}, \texttt{item\_name}, \texttt{game\_state\_balance}, \texttt{game\_state\_debt}, and \texttt{game\_state\_time\_left}. Logged events span purchases, investment deposits and returns, gambling bets and outcomes, credit card transactions, interest and bill payments, and collected rewards. These text events, together with the accompanying game-state snapshots, serve as the sole input to the assessment pipeline described in Section~\ref{sec:method}. Crucially, the event stream is captured transparently---players are never interrupted or aware that their actions are being recorded for assessment purposes---preserving the stealth assessment paradigm.

\section{Method}\label{sec:method}

\begin{figure*}[t]
\centering
\includegraphics[width=\textwidth]{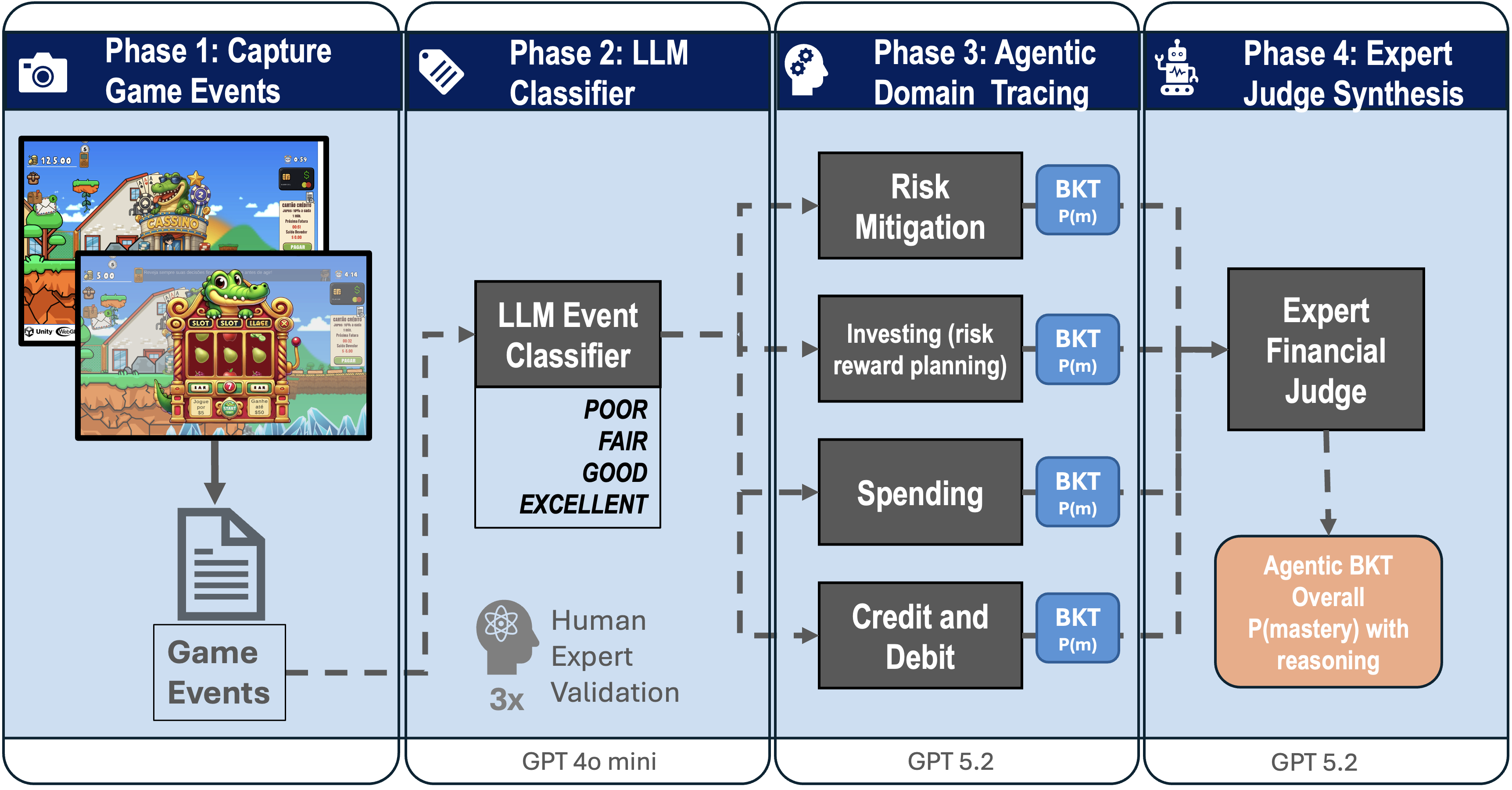}
\caption{Architecture of the Agentic BKT pipeline. Phase~1 captures structured game events. Phase~2 classifies each event on a four-point scale using GPT-4o~mini. Phase~3 routes classified events to four domain-specific GPT-5.2 agents, each producing a per-competency BKT mastery estimate. Phase~4 synthesizes domain-level scores into an overall $P(\text{mastery})$ via an expert judge agent. }
\label{fig:pipeline}
\end{figure*}

Figure~\ref{fig:pipeline} illustrates the full Agentic BKT pipeline. The architecture comprises four phases: (1)~event capture from gameplay (described in Section~III), (2)~LLM-based event classification, (3)~domain-specific agent assessment with per-competency BKT, and (4)~expert judge synthesis. The remainder of this section details Phases~2--4. We also describe the baseline single-LLM BKT approach and the pre/post-test instrument used for external validation. All system prompts used by the event classifier, the four domain agents, and the expert judge agent, together with the BKT implementation and analysis scripts, are openly available at the repository linked in the Data and Code Availability statement.

\subsection{Event-Level LLM Classification (Phase 2)}

Each game event captured in Phase~1 is independently classified by GPT-4o~mini into one of four ordinal levels: \textsc{poor}, \textsc{fair}, \textsc{good}, or \textsc{excellent}. The classifier receives the event description together with the full game-state context at the time of the action---including current balance, outstanding debt, time remaining, and the sequence of prior events in the session. Alongside the label, the model produces a reasoning trace that justifies the classification, providing interpretability for downstream agents.

To validate the LLM classifier, three independent domain experts with backgrounds in financial education classified the same set of events using the same four-point rubric. We report inter-rater reliability using Fleiss' $\kappa$ for the three human experts and for all four raters (three experts plus the LLM), as well as pairwise Cohen's $\kappa$ between each rater pair. As detailed in Section~\ref{sec:evaluation}, Fleiss' $\kappa$ among the three experts is 0.682 (substantial agreement), and including the LLM as a fourth rater yields Fleiss' $\kappa = 0.624$ (substantial), indicating that the automated classifier introduces only a marginal reduction in overall agreement.

\subsection{Domain-Specific Agent Assessment (Phase 3)}

Rather than treating all classified events as a single sequence, the Agentic BKT pipeline routes events to four domain-specific LLM agents, each specializing in one facet of the OECD/INFE competency framework:

\begin{itemize}
    \item \textbf{Risk Mitigation Agent}: evaluates gambling behavior, loss reactions, and risk-avoidance strategies.
    \item \textbf{Investing Agent}: assesses investment decisions including interest-rate selection, duration choices, and portfolio diversification.
    \item \textbf{Spending Agent}: analyzes purchasing patterns, price sensitivity, and budget management.
    \item \textbf{Credit \& Debt Agent}: examines credit card usage, debt accumulation, interest awareness, and repayment behavior.
\end{itemize}

Each agent is powered by GPT-5.2 and receives the \emph{full session} of events (not individual events in isolation), along with the Phase~2 classifications and reasoning traces. This session-level input is critical: it enables the agent to detect temporal trajectories that event-level classification cannot capture. For example, a player who gambles heavily in the first two minutes but stops after sustained losses and never returns to the casino demonstrates learning and risk awareness---a pattern the Risk Mitigation Agent can recognize and reward, whereas a purely event-level approach would assign the same negative labels to all early gambling events regardless of what followed.

Each domain agent outputs a binary sequence for BKT by mapping its assessment of each relevant event to 0 (\textsc{poor} or \textsc{fair}) or 1 (\textsc{good} or \textsc{excellent}). Standard BKT~\cite{corbett1995knowledge} is then run independently on each domain's binary sequence to estimate a per-competency $P(\text{mastery})$.

We fix the four BKT parameters to values drawn from the established BKT calibration literature rather than fitting them per participant. Specifically: $P(L_0) = 0.30$, a conservative prior reflecting that K-12 students typically arrive with limited but non-zero financial literacy; $P(T) = 0.10$, a moderate transition rate consistent with gradual skill acquisition observed in single-session gameplay~\cite{corbett1995knowledge}; $P(G) = 0.25$ and $P(S) = 0.10$, bounded to respect the identifiability constraints discussed by Beck and Chang~\cite{beck2007identifiability}, who recommend $P(G) \leq 0.30$ and $P(S) \leq 0.10$ to avoid the model degeneracy that occurs when guess and slip jointly approach the chance threshold. The asymmetric choice ($P(G) > P(S)$) reflects the open-ended nature of our gameplay: a low-mastery player can produce an event that the LLM classifier rates favorably (e.g., a single prudent purchase) more easily than a high-mastery player produces a poorly-rated one, making false positives more likely than false negatives. The decomposition into four parallel BKT models also ensures that mastery in one competency (e.g., responsible spending) does not inflate or mask deficits in another (e.g., risky gambling behavior).

We note that the fixed-parameter approach is consistent with prior BKT applications where labeled per-student fit data is unavailable~\cite{beck2007identifiability}. A full sensitivity analysis across the four parameters is left to future work, as our primary contribution is the architectural decomposition rather than the BKT calibration.

\subsection{Expert Judge Synthesis (Phase 4)}

The fourth phase mirrors the LLM-as-a-judge paradigm~\cite{zheng2023judging} but with domain-specific evidence. A GPT-5.2 judge agent receives all four per-competency BKT scores together with the full reasoning analyses produced by each domain agent. The judge synthesizes these inputs into an overall $P(\text{mastery})$ accompanied by a written justification. This design enables the judge to weigh domain competencies contextually---for instance, assigning greater weight to credit management when a student actively used the credit card system, or discounting a domain score when insufficient events were available for reliable estimation.

\subsection{Baseline: Single-LLM BKT}

To isolate the contribution of domain decomposition and expert synthesis, we compare the Agentic BKT pipeline against a single-LLM baseline that bypasses Phases~3 and~4. In this baseline, the Phase~2 event classifications are binarized directly (mapping \textsc{good}/\textsc{excellent} $\rightarrow$ 1, \textsc{poor}/\textsc{fair} $\rightarrow$ 0) and BKT is run on the entire undifferentiated sequence to produce a single $P(\text{mastery})$. This approach corresponds to a conventional LLM-enhanced BKT without competency decomposition.
We chose this baseline rather than DKT~\cite{piech2015deep}, AKT~\cite{ghosh2020context}, or LLMKT~\cite{scarlatos2025dialogue} because it functions as a \emph{controlled ablation} of our own pipeline: it shares Phase~2 with the Agentic pipeline and differs only in the absence of Phases~3 and~4, isolating the contribution of domain decomposition. DKT and AKT additionally require large labeled interaction corpora that our $N = 193$ deployment does not provide, and LLMKT was validated for tutoring dialogues rather than open-ended gameplay logs.

\subsection{Pre/Post-Test Instrument}

To obtain an external ground truth, participants completed a 15-question financial literacy test before and after gameplay. The test covers the five OECD/INFE competency domains exercised in the game, with three multiple-choice questions per domain assessing declarative and applied knowledge. Learning gain is defined as $\Delta = \text{post-test} - \text{pre-test}$, capturing within-subject knowledge change after gameplay. While the test follows the canonical five-domain OECD/INFE structure, the four agents in Phase~3 reflect a coarser decomposition aligned with the game’s behavioral mechanic clusters (Table~I).

\section{Evaluation}\label{sec:evaluation}

\subsection{Study Design}

We recruited 193 K-12 students from partnering schools. Each participant completed a pre-test, played the game for one or more sessions of 10--20 minutes, and then completed an identical post-test. Across all participants, 264 game sessions were recorded, yielding 15{,}447 total gameplay events (mean 58.5 events per session). Participants ranged from 7 to 12 years old (mean age = 9.8) and were recruited from one public school in Minas Gerais, Brazil.

A subset of participants completed more than one session, accounting for the difference between the 193 participants and 264 total sessions. To preserve a single-observation-per-participant design and avoid inflating statistical power through repeated measures, all correlational analyses (Tables~\ref{tab:correlations} and~\ref{tab:perdomain}) use each participant's \emph{first session lasting at least 10 minutes} as the unit of analysis ($N = 193$). We chose this criterion for two reasons. First, the first eligible session provides the fairest behavioral signal for assessing learning outcomes, as it captures the participant's initial engagement with the game free from carry-over effects of prior gameplay. Second, the 10-minute floor ensures that each analyzed session contains a sufficient volume of events across competency domains for reliable per-domain BKT estimation. All 193 participants had at least one session meeting this threshold. The full set of 264 sessions and 15{,}447 events is used only for descriptive statistics (Table~\ref{tab:eventdist}) and for the session-independent event-level rater agreement analysis (Table~\ref{tab:agreement}). Table~\ref{tab:eventdist} shows the distribution of events across financial literacy domains. Credit \& Debt events dominate (58.7\%), reflecting the prominence of the credit card mechanic, while Risk Mitigation (3.7\%) and Investing (2.3\%) are substantially sparser. The implications of this distributional imbalance for the per-domain analysis are discussed in Section~\ref{sec:discussion}. Parental consent was obtained for all participants, and the study protocol was approved by the institutional review board. Of the 193 participants, 71.5\% showed improvement on the post-test, 18.1\% declined, and 10.4\% were unchanged, yielding a mean learning gain of $\Delta = 2.01$ ($SD = 3.07$).

\begin{table}[t]
\centering
\caption{Distribution of gameplay events across financial literacy domains ($N = 15{,}447$ total events).}
\label{tab:eventdist}
\renewcommand{\arraystretch}{1.1}
\begin{tabular}{lrr}
\toprule
\textbf{Domain} & \textbf{Events} & \textbf{\%} \\
\midrule
Risk Mitigation & 576 & 3.7 \\
Investing & 354 & 2.3 \\
Spending & 4{,}401 & 28.5 \\
Credit \& Debt & 9{,}072 & 58.7 \\
Other / Unassigned & 1{,}044 & 6.8 \\
\midrule
\textbf{Total} & \textbf{15{,}447} & \textbf{100.0} \\
\bottomrule
\end{tabular}
\end{table}

\subsection{RQ1: LLM Classifier Agreement with Domain Experts}

Table~\ref{tab:agreement} reports inter-rater reliability for the four-point event classification rubric. Pairwise agreement between human experts (quadratic-weighted Cohen's $\kappa = 0.899$--$0.930$) is consistently high. The LLM classifier achieves $\kappa = 0.841$--$0.855$ against each expert---slightly below expert--expert pairs, but firmly in the \emph{almost perfect} range according to the Landis--Koch scale. At the multi-rater level, Fleiss' $\kappa$ drops only from 0.682 (three experts) to 0.624 (three experts + LLM), a reduction of 0.058 that preserves the \emph{substantial} agreement tier. These results confirm that the automated classifier operates within the range of human disagreement, validating its use as the first stage of the pipeline.

\begin{table}[t]
\centering
\caption{Pairwise weighted Cohen's $\kappa$ (quadratic) and exact agreement (\%) for event classification. Bottom rows show Fleiss' $\kappa$ for multi-rater agreement.}
\label{tab:agreement}
\renewcommand{\arraystretch}{1.1}
\begin{tabular}{lcc}
\toprule
\textbf{Comparison} & \textbf{$\kappa_w$ (quad.)} & \textbf{Exact \%} \\
\midrule
Expert 1 vs.\ Expert 2 & 0.899 & 74.1 \\
Expert 1 vs.\ Expert 3 & 0.930 & 78.9 \\
Expert 2 vs.\ Expert 3 & 0.912 & 77.8 \\
\midrule
LLM vs.\ Expert 1 & 0.841 & 70.2 \\
LLM vs.\ Expert 2 & 0.843 & 66.9 \\
LLM vs.\ Expert 3 & 0.855 & 68.9 \\
LLM vs.\ Majority Vote & 0.854 & 70.4 \\
\midrule
\multicolumn{2}{l}{Fleiss' $\kappa$ (3 experts)} & 0.682 \\
\multicolumn{2}{l}{Fleiss' $\kappa$ (3 experts + LLM)} & 0.624 \\
\bottomrule
\end{tabular}
\end{table}

\subsection{RQ2: Correlation with Learning Outcomes}

Table~\ref{tab:correlations} compares all three approaches---a random baseline, the single-LLM BKT baseline, and the Agentic BKT pipeline---against the external test measures. The random baseline, which assigns uniformly random mastery scores, produces no significant correlations with any outcome ($r = -0.075$, $p = 0.30$ for learning gain), confirming that the task is non-trivial.

The single-LLM BKT baseline yields a small positive trend with learning gain ($r = 0.095$, $p = 0.19$) and post-test ($r = 0.112$, $p = 0.12$), but neither reaches statistical significance. In contrast, the Agentic BKT pipeline achieves a significant correlation with learning gain ($r = 0.276$, $p = 0.0001$) and post-test score ($r = 0.333$, $p < 0.0001$), approximately tripling the effect size relative to the baseline. Crucially, neither approach correlates with pre-test scores (Agentic: $r = -0.036$, $p = 0.62$; Baseline: $r = -0.015$, $p = 0.83$), providing discriminant validity: the mastery estimates capture what was \emph{learned during gameplay}, not prior knowledge.

\begin{table}[t]
\centering
\caption{Pearson correlations ($r$) between mastery estimates and external test measures ($N = 193$). Significance: ${}^{***}p < .001$, ${}^{**}p < .01$, ${}^{*}p < .05$.}
\label{tab:correlations}
\renewcommand{\arraystretch}{1.15}
\begin{tabular}{lccc}
\toprule
\textbf{Approach} & \textbf{Learning Gain} & \textbf{Pre-test} & \textbf{Post-test} \\
\midrule
Random baseline & $-0.075$ & $\phantom{-}0.060$ & $-0.038$ \\
Single-LLM BKT & $\phantom{-}0.095$ & $-0.015$ & $\phantom{-}0.112$ \\
\textbf{Agentic BKT} & $\phantom{-}\mathbf{0.276}^{***}$ & $-0.036$ & $\phantom{-}\mathbf{0.333}^{***}$ \\
\bottomrule
\end{tabular}
\end{table}

\begin{figure}[t]
\centering
\includegraphics[width=\columnwidth]{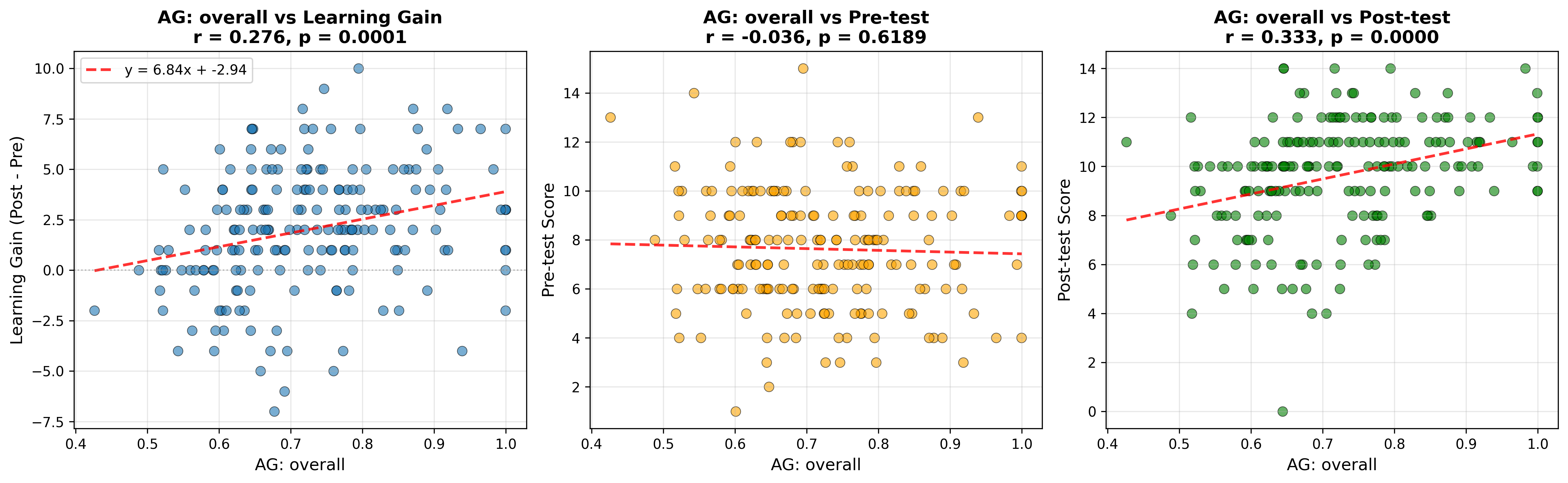}
\caption{Agentic BKT overall $P(\text{mastery})$ vs.\ learning gain (left), pre-test (center), and post-test (right). Significant correlations with learning gain ($r = 0.276$, $p = 0.0001$) and post-test ($r = 0.333$, $p < 0.0001$) demonstrate convergent validity, while the null pre-test correlation confirms discriminant validity.}
\label{fig:scatter_ag}
\end{figure}

\begin{figure}[t]
\centering
\includegraphics[width=\columnwidth]{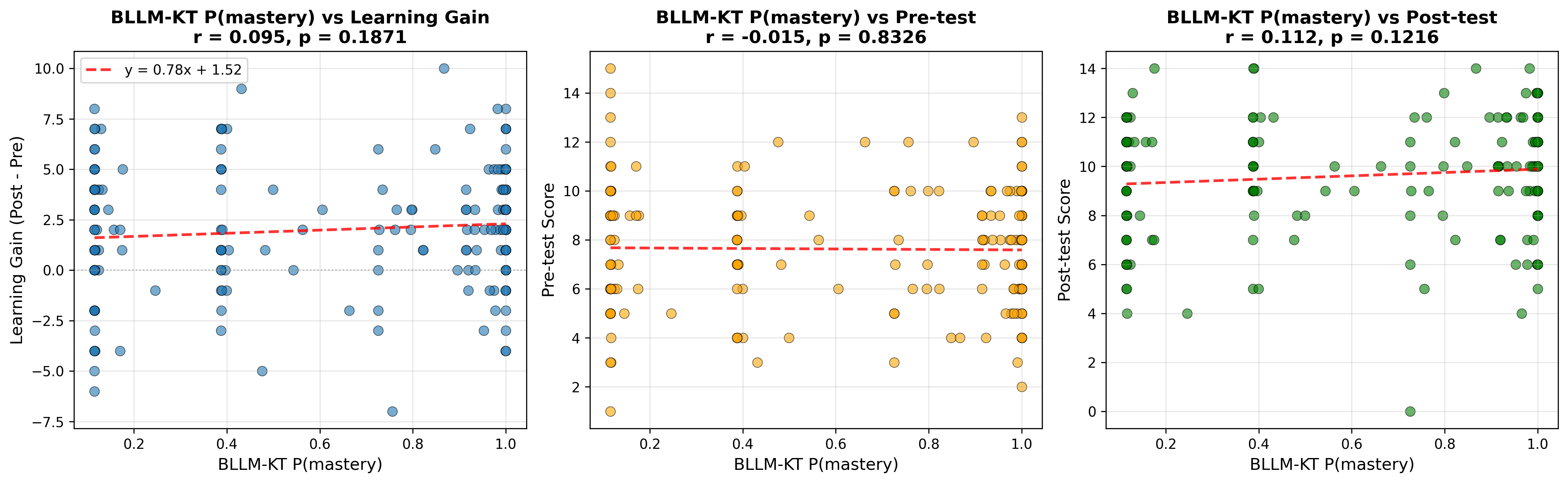}
\caption{Single-LLM BKT baseline $P(\text{mastery})$ vs.\ learning gain (left), pre-test (center), and post-test (right). No correlations reach significance, illustrating the limitation of event-level classification without domain decomposition.}
\label{fig:scatter_bllm}
\end{figure}

Figure~\ref{fig:scatter_ag} shows the Agentic BKT scatter plots against all three test measures, illustrating the significant positive trend with learning gain and post-test alongside the expected null relationship with pre-test. For comparison, the single-LLM baseline scatter plots are shown in Figure~\ref{fig:scatter_bllm}.

\subsection{RQ3: Contribution of Domain Decomposition}

To understand \emph{why} the multi-agent pipeline outperforms the monolithic baseline, Table~\ref{tab:perdomain} reports per-domain correlations with learning gain and post-test score. Credit management shows the strongest individual relationship with learning gain ($r = 0.204$, $p = 0.004$) and post-test ($r = 0.266$, $p = 0.0002$), followed closely by spending ($r = 0.197$, $p = 0.006$ and $r = 0.275$, $p = 0.0001$). Risk mitigation reaches significance for learning gain ($r = 0.159$, $p = 0.027$) but not post-test ($r = 0.113$, $p = 0.12$). Investment logic trends positive but does not reach significance for either outcome ($r = 0.109$, $p = 0.13$); we note that this domain exhibits very low variance ($SD = 0.036$, with a ceiling near 1.0), likely reflecting that most players made broadly reasonable investment choices, limiting the agent's discriminative power.

These per-domain results support two conclusions. First, no single competency accounts for the overall pipeline's predictive validity---rather, the judge agent's synthesis across domains captures a richer signal than any individual component. Second, the decomposition itself is informative: credit and spending behaviors appear to be the strongest behavioral indicators of financial literacy learning in this game, a finding that can directly inform future game design.

\begin{table}[t]
\centering
\caption{Per-domain agent correlations with external test measures ($N = 193$). Significance: ${}^{***}p < .001$, ${}^{**}p < .01$, ${}^{*}p < .05$.}
\label{tab:perdomain}
\renewcommand{\arraystretch}{1.15}
\begin{tabular}{lcc}
\toprule
\textbf{Domain Agent} & \textbf{Learning Gain} & \textbf{Post-test} \\
\midrule
Risk Mitigation & $0.159^{*}$ & $0.113$ \\
Investing & $0.109$ & $0.069$ \\
Spending & $0.197^{**}$ & $0.275^{***}$ \\
Credit \& Debt & $0.204^{**}$ & $0.266^{***}$ \\
\midrule
\textbf{Overall (Judge)} & $\mathbf{0.276}^{***}$ & $\mathbf{0.333}^{***}$ \\
\bottomrule
\end{tabular}
\end{table}

No domain agent correlates significantly with pre-test scores (all $|r| < 0.10$, $p > 0.18$), confirming that the discriminant validity observed at the overall level holds consistently across individual competencies.

\section{Discussion}
\label{sec:discussion}

The central finding of this work is that decomposing financial literacy assessment into domain-specific agents approximately triples the correlation between gameplay-derived mastery estimates and external learning outcomes ($r = 0.095 \rightarrow 0.276$). This improvement is not merely statistical: it reflects a substantive architectural insight about how financial competence should be modeled. The three experimental conditions---random baseline, single-LLM BKT, and Agentic BKT---form a progressive ablation: the random baseline validates that the task is non-trivial; the single-LLM baseline isolates the contribution of LLM event classification (Phase~2); and the full Agentic pipeline adds domain decomposition and expert synthesis (Phases~3--4). The jump in predictive validity occurs entirely at the domain-decomposition stage, pinpointing it as the architectural component responsible for the performance gain. We discuss the implications, situate the effect sizes in context, and acknowledge limitations.

\subsection{Why Domain Decomposition Matters}

The single-LLM baseline treats every gameplay event as an interchangeable observation drawn from one latent skill. In practice, however, a student who spends recklessly but invests wisely possesses a very different competency profile from one who exhibits the reverse pattern---yet both may produce similar aggregate BKT trajectories when events are pooled. By routing events to specialized agents, the Agentic pipeline preserves these distinctions. Moreover, domain agents reason over \emph{entire sessions}, enabling them to detect temporal trajectories that event-level classification cannot capture. Consider a player who gambles heavily in the first two minutes, loses 30 coins, and then avoids the casino for the remainder of the session. The event-level classifier assigns \textsc{poor} to each early gambling event regardless of what follows; the Risk Mitigation Agent, by contrast, recognizes the behavioral arc as evidence of learning and adjusts its assessment upward. This session-level reasoning is the mechanism through which the multi-agent approach extracts signal that a monolithic model discards.

A concrete example illustrates this advantage. Player~68, a nine-year-old participant, began the session by gambling on the slot machine, losing coins across several spins. The event-level classifier correctly labeled these early bets as \textsc{poor}. However, the player then shifted strategy entirely---investing all available coins at 5\% interest, collecting earnings across multiple payout cycles, and making only small purchases with surplus funds. No further gambling occurred for the remainder of the session. The Risk Mitigation Agent recognized this behavioral trajectory as evidence of learning: early experimentation followed by a decisive pivot away from risk. The single-LLM baseline, by contrast, produced a moderate overall score diluted by the early \textsc{poor} labels. In a post-session interview, this participant spontaneously stated: ``I learned that gambling sucks, because I lost my money there \ldots\ investing is a better option''---and added that they intended to advise a family member to stop gambling. This qualitative evidence aligns precisely with the Risk Mitigation Agent's assessment and illustrates the kind of within-session learning trajectory that domain-specific, session-level reasoning is designed to capture.

\subsection{Interpreting Effect Sizes}

Correlations of $r = 0.276$ (learning gain) and $r = 0.333$ (post-test) are moderate by conventional standards, but several factors argue that these are meaningful in context. First, stealth assessment from open-ended gameplay is fundamentally harder than structured assessment: there are no explicit questions, no constrained response formats, and no guarantee that a given session will exercise all competency domains. Second, the pre/post-test measures declarative knowledge, while gameplay captures procedural and behavioral competence---constructs that overlap but are not identical. Perfect correlation between these two measurement modalities would be surprising and potentially suspicious. Third, the near-zero correlations with pre-test scores ($r = -0.036$, $p = 0.62$) provide discriminant validity: the pipeline measures what was \emph{learned during gameplay}, not prior knowledge that students brought to the session.

\subsection{Implications for Game Design}

The per-competency breakdown in Table~\ref{tab:perdomain} reveals that credit management and spending are the strongest behavioral predictors of learning gain, while investment logic exhibits a ceiling effect. Table~\ref{tab:eventdist} suggests a contributing factor: investing accounts for only 2.3\% of all events, compared to 58.7\% for credit and debt. Future game versions could introduce more varied investment scenarios---competing interest rates, penalty clauses, or time-pressure decisions---to increase behavioral diversity in that domain. This event imbalance also has an interpretive consequence: the per-domain correlations in Table~\ref{tab:perdomain} should be read in light of the volume of evidence each agent receives, rather than as direct comparisons of intrinsic difficulty across competencies.

More broadly, the Agentic BKT pipeline offers game designers a diagnostic tool: by examining which domain agents produce significant correlations with learning outcomes, designers can identify where a game succeeds or falls short in exercising target competencies, enabling evidence-driven iteration without a full redesign-and-retest cycle. The architecture is in principle domain-agnostic, though its applicability to other serious-game domains such as sustainability, digital citizenship, or health literacy remains to be empirically tested.

\subsection{Limitations}

Several limitations should be acknowledged. Although the sample ($N = 193$) was sufficient to detect the observed effects, it may not capture the full diversity of financial literacy behaviors, especially in older populations. Results are based on a single game, limiting generalizability to other serious games. The pipeline relies on commercial LLMs (GPT-4o mini and GPT-5.2), raising reproducibility and cost concerns; future work should explore open-source alternatives. While BKT was selected for its interpretability, more expressive approaches such as DKT may improve competency estimation. In addition, the multi-agent inference pipeline currently operates only as a post-session analysis tool due to latency and cost constraints. Processing one session requires approximately 93 API calls, costing about \$0.048 per student (\$9.26 for the full experiment), which remains practical for research but may require optimization for large-scale deployment. Finally, our evaluation isolates the contribution of domain decomposition through a single-LLM ablation; future work should compare Agentic BKT against alternative KT approaches such as DKT, AKT, and LLMKT on the same gameplay data.

\section{Conclusion}

We introduced the Agentic BKT pipeline, a multi-agent LLM framework for assessing financial literacy from open-ended serious game events through event classification, domain-specific reasoning aligned with the OECD/INFE framework, and expert synthesis. Evaluated with 193 K-12 participants, the pipeline produced mastery estimates significantly correlated with learning gain ($r = 0.276$, $p = 0.0001$) and post-test scores ($r = 0.333$, $p < 0.0001$), while remaining uncorrelated with prior knowledge. Compared to a single-LLM baseline, the approach approximately tripled predictive validity and provided evidence of both convergent and discriminant validity.

Future work will focus on: (1) reducing inference latency to support real-time adaptive gameplay, (2) validating the framework across broader populations and game genres, and (3) exploring open-source LLMs to improve reproducibility and reduce cost. More broadly, the Agentic BKT framework is in principle domain-agnostic, with applicability to other serious games grounded in established competency frameworks remaining to be empirically tested.

\smallskip
\noindent\textbf{Data and Code Availability.} Gameplay logs and pre/post-test scores cannot be publicly released due to ethical restrictions involving minor participants. To support reproducibility, all LLM prompts, the BKT implementation, and the analysis scripts are publicly available at: \url{https://github.com/gabrielmsantos/LAKT}.

\smallskip
\noindent\textbf{Acknowledgements.}
We thank Escola Estadual Messias Pedreiro (Uberlândia, Brazil) for supporting the validation of our game-based educational system; ADASYS Game Club (Federal University of Uberlândia) for assistance during deployment at the partner school; Diogo Abílio de Alcantara for the game design; and Professor Stéphane Julia for coordinating the related outreach activities. This work was supported by the National Council for Scientific and Technological Development (CNPq), Grant \#302833/2025-0.

\bibliographystyle{IEEEtran}
\bibliography{references}

\begin{thebibliography}{10}
\providecommand{\url}[1]{#1}
\csname url@samestyle\endcsname
\providecommand{\newblock}{\relax}
\providecommand{\bibinfo}[2]{#2}
\providecommand{\BIBentrySTDinterwordspacing}{\spaceskip=0pt\relax}
\providecommand{\BIBentryALTinterwordstretchfactor}{4}
\providecommand{\BIBentryALTinterwordspacing}{\spaceskip=\fontdimen2\font plus
\BIBentryALTinterwordstretchfactor\fontdimen3\font minus \fontdimen4\font\relax}
\providecommand{\BIBforeignlanguage}[2]{{%
\expandafter\ifx\csname l@#1\endcsname\relax
\typeout{** WARNING: IEEEtran.bst: No hyphenation pattern has been}%
\typeout{** loaded for the language `#1'. Using the pattern for}%
\typeout{** the default language instead.}%
\else
\language=\csname l@#1\endcsname
\fi
#2}}
\providecommand{\BIBdecl}{\relax}
\BIBdecl

\bibitem{oecd2020infe}
\BIBentryALTinterwordspacing
{OECD}, ``{OECD/INFE} 2020 international survey of adult financial literacy,'' OECD, Paris, Tech. Rep., 2020. [Online]. Available: \url{https://www.oecd.org/financial/education/oecd-infe-2020-international-survey-of-adult-financial-literacy.pdf}
\BIBentrySTDinterwordspacing

\bibitem{oecd2024pisa}
------, \emph{{PISA} 2022 Results (Volume {IV}): How Financially Smart Are Students?}, ser. PISA.\hskip 1em plus 0.5em minus 0.4em\relax Paris: OECD Publishing, 2024.

\bibitem{amagir2018review}
A.~Amagir, W.~Groot, H.~M. van~den Brink, and A.~Wilschut, ``A review of financial-literacy education programs for children and adolescents,'' \emph{Citizenship, Social and Economics Education}, vol.~17, no.~1, pp. 56--80, 2018.

\bibitem{aprea2018instructional}
C.~Aprea, J.~Schultheis, and K.~Stolle, \emph{Instructional Integration of Digital Learning Games in Financial Literacy Education}, 11 2017.

\bibitem{pfandler2021happy}
A.~M. Pf\"{a}ndler, ``Development and pilot testing of a financial literacy game for young adults: The {Happy Life Game},'' in \emph{Game-based Learning Across the Disciplines}, ser. Advances in Game-Based Learning, C.~Aprea and D.~Ifenthaler, Eds.\hskip 1em plus 0.5em minus 0.4em\relax Cham: Springer, 2021, pp. 59--84.

\bibitem{platz2025financial}
J.~Platz \emph{et~al.}, ``Financial literacy games---increasing utility value by instructional design in upper secondary education,'' \emph{Education Sciences}, vol.~15, no.~2, p. 227, 2025.

\bibitem{shute2009melding}
V.~J. Shute, M.~Ventura, M.~I. Bauer, and D.~Zapata-Rivera, ``Melding the power of serious games and embedded assessment to monitor and foster learning: Flow and grow,'' in \emph{Serious Games: Mechanisms and Effects}, U.~Ritterfeld, M.~Cody, and P.~Vorderer, Eds.\hskip 1em plus 0.5em minus 0.4em\relax Routledge, 2009, pp. 295--321.

\bibitem{shute2011stealth}
V.~J. Shute, ``Stealth assessment in computer-based games to support learning,'' in \emph{Computer Games and Instruction}, S.~Tobias and J.~D. Fletcher, Eds.\hskip 1em plus 0.5em minus 0.4em\relax Charlotte, NC: Information Age Publishers, 2011, pp. 503--524.

\bibitem{shute2013measuring}
V.~J. Shute and M.~Ventura, \emph{Measuring and Supporting Learning in Games: Stealth Assessment}.\hskip 1em plus 0.5em minus 0.4em\relax Cambridge, MA: MIT Press, 2013.

\bibitem{corbett1995knowledge}
A.~T. Corbett and J.~R. Anderson, ``Knowledge tracing: Modeling the acquisition of procedural knowledge,'' \emph{User Modeling and User-Adapted Interaction}, vol.~4, no.~4, pp. 253--278, 1995.

\bibitem{schmucker2024towards}
\BIBentryALTinterwordspacing
R.~Schmucker \emph{et~al.}, ``Towards modeling learner performance with large language models,'' \emph{arXiv preprint arXiv:2403.14661}, 2024. [Online]. Available: \url{https://arxiv.org/abs/2403.14661}
\BIBentrySTDinterwordspacing

\bibitem{scarlatos2025dialogue}
A.~Scarlatos, R.~S. Baker, and A.~Lan, ``Exploring knowledge tracing in tutor-student dialogues using {LLMs},'' in \emph{Proceedings of the 15th International Learning Analytics and Knowledge Conference (LAK 2025)}.\hskip 1em plus 0.5em minus 0.4em\relax Dublin, Ireland: ACM, 2025.

\bibitem{llmagents2025education}
\BIBentryALTinterwordspacing
X.~Wang \emph{et~al.}, ``{LLM} agents for education: Advances and applications,'' \emph{arXiv preprint arXiv:2503.11733}, 2025. [Online]. Available: \url{https://arxiv.org/abs/2503.11733}
\BIBentrySTDinterwordspacing

\bibitem{corbin2022nova}
\BIBentryALTinterwordspacing
J.~Corbin, J.~Dalton, R.~Bhatt, and D.~Ariely, ``A behavioral science approach to financial literacy games,'' Center for Advanced Hindsight, Duke University / GBH NOVA, 2022. [Online]. Available: \url{https://advanced-hindsight.com/wp-content/uploads/2022/03/CAH-NOVA.pdf}
\BIBentrySTDinterwordspacing

\bibitem{rasco2020fincraft}
\BIBentryALTinterwordspacing
A.~Rasco, J.~Chan, G.~Peko, and D.~Sundaram, ``{FinCraft}: Immersive personalised persuasive serious games for financial literacy among young decision-makers,'' in \emph{Proceedings of the 53rd Hawaii International Conference on System Sciences (HICSS)}, 2020, pp. 1--10. [Online]. Available: \url{https://scholarspace.manoa.hawaii.edu/items/aeae2621-8ab3-4d76-b73c-ecdcfdbbe563}
\BIBentrySTDinterwordspacing

\bibitem{piech2015deep}
C.~Piech, J.~Bassen, J.~Huang, S.~Ganguli, M.~Sahami, L.~Guibas, and J.~Sohl-Dickstein, ``Deep knowledge tracing,'' in \emph{Advances in Neural Information Processing Systems}, vol.~28.\hskip 1em plus 0.5em minus 0.4em\relax Curran Associates, Inc., 2015, pp. 505--513.

\bibitem{ghosh2020context}
A.~Ghosh, N.~Heffernan, and A.~S. Lan, ``Context-aware attentive knowledge tracing,'' in \emph{Proceedings of the 26th ACM SIGKDD International Conference on Knowledge Discovery and Data Mining}.\hskip 1em plus 0.5em minus 0.4em\relax ACM, 2020, pp. 2330--2339.

\bibitem{norris2025ntkt}
\BIBentryALTinterwordspacing
M.~Norris and S.~Bulathwela, ``Next token knowledge tracing: Exploiting pretrained {LLM} representations to decode student behaviour,'' \emph{arXiv preprint arXiv:2511.02599}, 2025. [Online]. Available: \url{https://arxiv.org/abs/2511.02599}
\BIBentrySTDinterwordspacing

\bibitem{li2025twotkt}
L.~Li, Z.~Wang, J.~M. Jose, and X.~Ge, ``Llm supporting knowledge tracing leveraging global subject and student specific knowledge graphs,'' \emph{Information Fusion}, p. 103577, 2025.

\bibitem{zheng2023judging}
L.~Zheng, W.-L. Chiang, Y.~Sheng, S.~Zhuang, Z.~Wu, Y.~Zhuang, Z.~Lin, Z.~Li, D.~Li, E.~P. Xing, H.~Zhang, J.~E. Gonzalez, and I.~Stoica, ``Judging {LLM}-as-a-judge with {MT-Bench} and {Chatbot Arena},'' in \emph{Advances in Neural Information Processing Systems}, vol.~36.\hskip 1em plus 0.5em minus 0.4em\relax Curran Associates, Inc., 2023.

\bibitem{beck2007identifiability}
J.~E. Beck and K.-m. Chang, ``Identifiability: A fundamental problem of student modeling,'' in \emph{Proc. 11th Int. Conf. on User Modeling (UM)}, 2007, pp. 137--146.

\end{thebibliography}

\end{document}